\documentclass[sigconf, nonacm]{acmart}
\newcommand\vldbyear{2025}
\newcommand\vldbworkshop{Tabular Data Analysis (TaDA)}
\newcommand\vldbauthors{\authors}
\newcommand\vldbtitle{\shorttitle} 
\newcommand\vldbavailabilityurl{https://github.com/panagiotis-koletsis/cpa4}
\newcommand\vldbpagestyle{plain}

\usepackage{algorithm}
\usepackage{algorithmic}
\usepackage{arydshln}

\begin{document}
\title{Relationship Detection on Tabular Data Using Statistical Analysis and Large Language Models}

\author{Panagiotis Koletsis}
\affiliation{%
  \institution{Harokopio University}
  \city{Athens}
  \country{Greece}
}
\email{pkoletsis@hua.gr}

\author{Christos Panagiotopoulos}
\affiliation{%
  \institution{Harokopio University}
  \city{Athens}
  \country{Greece}
}
\email{chris.panagiotop@hua.gr}

\author{Georgios Th. Papadopoulos}
\affiliation{%
  \institution{Harokopio University}
  \city{Athens}
  \country{Greece}
}
\email{g.th.papadopoulos@hua.gr}

\author{Vasilis Efthymiou}
\affiliation{%
  \institution{Harokopio University}
  \streetaddress{}
  \city{Athens}
  \country{Greece}
}
\email{vefthym@hua.gr}



\begin{abstract}
Over the past few years, table interpretation tasks have made significant progress due to their importance and the introduction of new technologies and benchmarks in the field. This  work experiments with a hybrid approach for detecting relationships among columns of unlabeled tabular data, using a Knowledge Graph (KG) as a reference point, a task known as CPA. This approach leverages large language models (LLMs) while employing statistical analysis to reduce the search space of potential KG relations. The main modules of this approach for reducing the search space are domain and range constraints detection, as well as relation co-appearance analysis. The experimental evaluation on two benchmark datasets provided by the SemTab challenge assesses the influence of each module and the effectiveness of different state-of-the-art LLMs at various levels of quantization. The experiments were performed, as well as at different prompting techniques. The proposed methodology, which is publicly available on github, proved to be competitive with state-of-the-art approaches on these datasets.
\end{abstract}

\maketitle

\pagestyle{\vldbpagestyle}
\begingroup\small\noindent\raggedright\textbf{VLDB Workshop Reference Format:}\\
\vldbauthors. \vldbtitle. VLDB \vldbyear\ Workshop: \vldbworkshop.\\ 
\endgroup
\begingroup
\renewcommand\thefootnote{}\footnote{\noindent
This work is licensed under the Creative Commons BY-NC-ND 4.0 International License. Visit \url{https://creativecommons.org/licenses/by-nc-nd/4.0/} to view a copy of this license. For any use beyond those covered by this license, obtain permission by emailing \href{mailto:info@vldb.org}{info@vldb.org}. Copyright is held by the owner/author(s). Publication rights licensed to the VLDB Endowment. \\
\raggedright Proceedings of the VLDB Endowment. 
ISSN 2150-8097. \\
}\addtocounter{footnote}{-1}\endgroup

\ifdefempty{\vldbavailabilityurl}{}{
\vspace{.3cm}
\begingroup\small\noindent\raggedright\textbf{PVLDB Artifact Availability:}\\
The source code, data, and/or other artifacts have been made available at \url{\vldbavailabilityurl}.
\endgroup
}

\section{Introduction}
Understanding the structure, semantics, and contents of tabular data is a foundational task in data integration, information retrieval, and Knowledge Graph (KG) population. An important step in this process is determining how columns in a table relate to one another. This problem is already difficult when column headers exist in a table (due to data heterogeneity), but it becomes particularly challenging, even for human experts, when the table lacks meaningful headers. The Column Property Annotation (CPA) task of the SemTab challenge~\cite{semtab} addresses this problem by aiming to identify semantic relationships between columns, such as ``authorOf'' or ``locatedIn'', as defined in a reference KG. This task is especially challenging in practice, due to the absence of informative column names, the broad and overlapping semantics of KG properties, and the large search space of potential relationships (e.g., there are roughly 3,000 properties in DBpedia and 1,500 in Schema.org).

Semantic table interpretation approaches that rely on Large language models (LLMs) have been recently proposed, with promising results~\cite{DBLP:conf/semweb/DasoulasYDD23,DBLP:conf/semtab/BikimYJOR0A24,DBLP:conf/semweb/HuynhCLLT22}. However, LLMs are also prone to hallucinations and the so-called knowledge injection has been suggested as a means of countering this problem~\cite{DBLP:conf/esws/MartinoIT23}. Therefore, in this preliminary work, we explore the use of data analysis to detect constraints that will reduce the search space of candidate relationships between columns, this way reducing the chances of errors, before prompting an LLM for the CPA task. Specifically, we detect domain and range restrictions of candidate relationships from the target KG, which, in our case is Schema.org, and combine them with domain and range restrictions detected in an input table. Moreover, we extract co-appearance statistics from the training set and exploit them to incrementally reduce the candidate relationships found in a table, as we detect some first relationships for that table. Finally, we explore several LLMs and prompting techniques and evaluate their performance.

In summary, the contributions of this work are the following:
\begin{itemize}
  \item A hybrid approach for CPA incorporating LLMs and statistical analysis.
  \item An experimental evaluation of several low-billion-parameter LLMs, with different quantization levels and prompting techniques.
  \item Ablation studies to define the impact of each of the modules of our hybrid approach.
  \item An open-source solution to CPA that is publicly available on \url{\vldbavailabilityurl}.

\end{itemize}

\section{Related Work}
The challenge of understanding improperly stored data has been recognized for many decades, leading to the exploration of new solutions, mostly based on statistics and machine learning ~\cite{DBLP:journals/ijdar/ZanibbiBC04,DBLP:conf/www/WangH02}, as well as lookup-based and LLM-based approaches~\cite{DBLP:journals/corr/abs-2411-11891}. 

\textbf{Statistical-based approaches.}
One of the foundational works in this area~\cite{DBLP:journals/pvldb/CafarellaHWWZ08} tackled the following problems: schema auto-complete, synonym finding and joining schemas by statistical approach. Schema auto-complete was possible by incorporating co-appearance and accounting probabilities, for example, when the user enters a make, the system suggests model, year, price, mileage, and color. Synonym finding was possible by knowing that synonyms would never co-appear in the same schema and that they would have similar content. Finally, they joined schemas by clustering similar ones. Our co-appearance statistics are inspired by this work. 

\textbf{Look-up based approaches.}
MTab~\cite{DBLP:conf/semweb/NguyenKI019a}, a work that dominated the SemTab challenge in its first few years, utilized entity look up by searching across local indices built on the target KG (e.g., DBpedia). Additionally, MTab introduced a literal matching approach to align table cell values with the corresponding properties of KGs~\cite{DBLP:conf/semweb/NguyenKI019a}. DAGOBAH~\cite{DBLP:conf/semweb/HuynhLCDLMT21}, a system that later dominated the SemTab challenge, further explored the above logic introducing preprocessing steps such as identification of orientation detection, header detection, key column detection and column primitive typing while also introducing a pre-scoring algorithm. Some of those ideas were introduced in works that preceded SemTab (e.g.,~\cite{DBLP:conf/semweb/EfthymiouHRC17,DBLP:conf/wims/RitzeLB15,DBLP:conf/webdb/LehmbergB16}).   

\textbf{LLM based.}
With all the advancements of LLMs in the past few years, researchers explored their capabilities in table interpretation tasks. For instance,  ~\cite{DBLP:conf/semtab/BikimYJOR0A24}  used a GPT3 model incorporating few-shot and zero-shot learning for solving SemTab tasks. More advanced works further explore even the fine tuning of LLMs for table interpretation tasks ~\cite{DBLP:journals/corr/abs-2310-09263} following even a foundational model approach for table interpretation ~\cite{DBLP:conf/naacl/ZhangYL024}.

\textbf{Hybrid.}
Interestingly, some hybrid works try to leverage the privileges of the aforementioned approaches, combining, for example, LLMs and look-up based techniques ~\cite{DBLP:conf/semweb/HuynhCLLT22}.
TorchicTab~\cite{DBLP:conf/semweb/DasoulasYDD23} employs a dual-system approach for semantic table annotation, combining heuristics with classification. The heuristic component, TorchicTab-Heuristic, uses RDF graphs like Wikidata for candidate lookup and ranking via multiple search strategies and semantic similarity, handling diverse table structures through contextual analysis and majority voting. The classification component, TorchicTab-Classification, treats column type and property annotation as a multi-class task using the DODUO language model (a finetuned BERT ~\cite{DBLP:conf/naacl/DevlinCLT19}) and sub-table sampling to manage token limits and preserve context. This combined approach achieved top performance in the SemTab 2023 challenge. 

\section{Methodology}
\begin{figure}
  \centering
  \includegraphics[width=\linewidth]{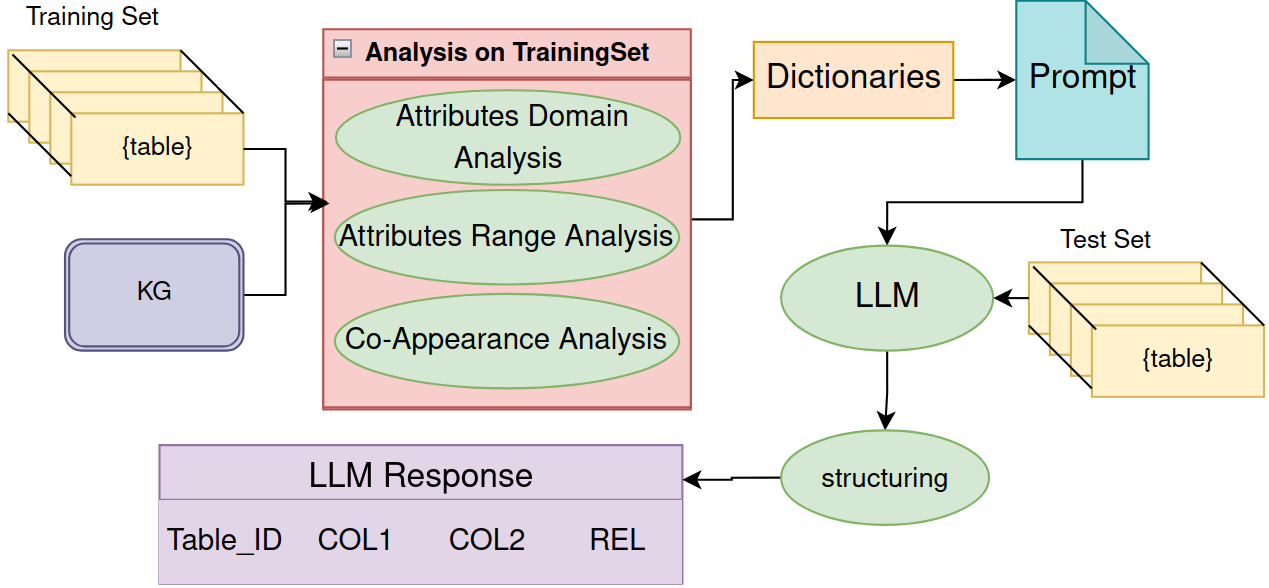}
  \caption{Methodology Overview.}
  \label{fig:MethodologyOverview}
\end{figure}

In this section, we describe our methodology, following the overview presented in Figure~\ref{fig:MethodologyOverview}.

\textbf{Base.}
In the base architecture, 
the table, the column of interest, and the complete list of possible relations are parsed into the template and subsequently processed by the LLM (skipping the Analysis and Dictionaries of Figure~\ref{fig:MethodologyOverview}). If the output does not match the desired format (i.e., a single word from the relation list), a simple yet effective technique is applied by parsing the output of the first prompt and asking for a single word of interest. If this fails again, then we just re-ask a different model with the initial prompt.


\textbf{Domain Attributes Analysis.}
Rather than passing the entire list of possible relations found in  schema.org, the domain attribute analysis focuses on identifying a smaller, more relevant set of relations for each entity type. This is possible because each data point in the training set table starts with the corresponding domain. In order to achieve our goal, we scan the training set and append, in a dictionary containing all the property types, the corresponding relation to the corresponding key on the dictionary.

\textbf{Range Attributes Analysis.}
Building on the initial idea, we further reduce the search space of relations by filtering them based on their primitive type. The proposed module for attribute range analysis scans the training set and reads the first element of the table for the corresponding column in the data point, then classifies it into one out of 
four types - String, Number, Date, and URL - using parsers and regular expressions.
The produced dictionary has one major drawback: due to inconsistencies in the dataset and the date parsing process, it contains some outliers. For example, some relations may appear in more than 2-3 primitive types. In order to improve this module, we incorporate a threshold for excluding rarely used relations. This threshold discarded all the relationships appearing less frequently than 5 \% of the most frequent one. That way, some outliers contained by the dataset are not accounted by the LLM and therefore, do not add noise to the LLM prompt.


\textbf{Co-appearance Analysis.} The idea behind the co-appearance module is that relationships that do not co-appear in the same table in the training set are not very likely to co-appear in the same table in the test set, either.
For each relationship of a specific domain, we build a corresponding dictionary (e.g., we build a dictionary for the property Book.name and another one for Person.name). This dictionary will contain the other relationship that co-occur with the initial relation in the same table, hopefully further narrowing the search space. In addition, while parsing the table's columns, the LLM's predictions for the previous columns are also removed from the candidate set of the next columns, since we assume that a relation only appears once per table. This technique has one major drawback: in order to be successful, the initial prediction of the model must be correct. Otherwise, the progressively smaller search space of candidate relations for the next column will be incorrect, propagating the errors of previous detections and
asking the model to choose a relation from a list that does not contain the correct one. In this preliminary approach, we scan columns from left to write, although other approaches (e.g., from the most likely to be annotated correctly to the least likely) are also worth exploring.



\textbf{Inference phase.} During the inference phase, the domain of the table is first extracted (i.e., the Table Topic Detection - TD - task of SemTab). Then, for each column in the ground truth, its type is identified by the aforementioned process (R). Finally, the intersection of these two lists is provided in the prompt template. 
It is important to mention that (a) we take a sample of 500 rows from each table for our analyses, and that (b) the provided script runs dynamically and produces all the dictionary without any manual effort.
This step takes place offline with a separate script, since it only makes sense to run once per dataset.


\section{Experiments}

In this section, we describe the experimental setting and provide and analyze the experimental results of our study. 

\subsection{Experimental Setting}

\textbf{Hardware-Software.} All the experiments were performed on a single GPU (RTX 3080 Ti with 12GB VRAM). The access to the models was achieved through Ollama (v.0.6.6) and Langchain.

\textbf{Models.} 
The focus was primary with the latest models from Ollama ranging from 12 to 72 billion parameters and from 2 bit to 4 bit quantization. Models such as DeepSeek-r1, Gemma3, phi4, Llamma3.3 are tested and \texttt{Qwen2.5:32B-Instruct-Q3\_K\_L} was evaluated in greater detail.


\textbf{Datasets.} In our experiments, we employ the Round~1 and Round~2 SemTab 2023 SOTAB~\cite{DBLP:conf/semweb/KoriniPB22} datasets for the CPA task, with schema.org as their target KG.
The Round 1 dataset consist of 50 predefined relations originating from 7 different domains. The Round 2 dataset follows the exact same format as the R1 dataset, but it consists of 105 relations from 17 distinct domains. In addition, it contains almost three times the size of data points compared to R1.



\textbf{Evaluation metrics.} We employ the standard metrics used in the evaluation of SemTab, i.e., Precision, Recall, Macro\_F1, and Micro\_F1, as well as execution time in seconds. To ensure that all methods are evaluated consistently, the SemTab challenge provides an evaluator Python script. 



\textbf{Base.}
Different LLMs with varying levels of quantization were tested. As shown in \autoref{tab:R1}, the Qwen family of models demonstrated superior results. This was likely due to the fact that Qwen models were specifically trained to understand and produce structured formats. Qwen-32B had fewer failed iterations, indicating better formatted output. Unexpectedly, LLaMA 3.3 failed to be competitive despite its larger size. Qwen2.5-72B Q2 showed inconsistent results compared to Qwen2.5-32B Q3 and had significantly more failed iterations, suggesting that distillation affected its ability to handle structured output. The other models were not competitive and exhibited increased execution time due to the activation of the structuring component.

\textbf{Domain (D).} In both datasets (\autoref{tab:R1} and \autoref{tab:R2}), domain extraction showed the most significant and clear improvements, indicating that a reduced and more relevant search space helps the LLM perform better. Additionally, both datasets exhibited a lower failure ratio and execution time. In the case of R1, a 2.5\% reduction in execution time per iteration was observed compared to base. This suggests that fewer choices led to more efficient decision-making by the model. In the R1 dataset, the Micro\_F1 score increased by 16.7\%, while in R2, the same metric improved by 17.3\%. 

\textbf{Range (R).} On its own on the R1 dataset, it increased all the metrics spotted compared to base. Most notably the precision was increased by 6 \%.  In the R2 dataset, all observed metrics declined. However, similar to the Domain approach, a reduction in execution time was noted on both datasets.

\textbf{Co-Appearance (C).} When used alone, it had a notably negative impact, as any initial error made by the LLM was propagated through subsequent iterations. The results confirm this behavior, showing a significant drop in performance metrics and a substantial increase in the failure rate. 

\textbf{Range \& Domain (RD).} The combination of type and domain analysis yielded the best results in both datasets, despite the fact that type analysis alone had a mixed impact. In the R1 dataset, Micro\_F1 increased by an additional 1.2\% compared to using domain analysis alone, and by 18.2\% compared to the base. In the R2 dataset, a further improvement by 5\% in Micro\_F1 was observed, along with a 6.5\% increase in precision. Notably, micro\_f1 improved by 23\% compared to the base. Although the proposed method didn't achieve SOTA performance, it has a major advantage over TorchicTab and MUT2KG. This advantage is that it does not require any finetuning unlike the above mentioned methods.

\textbf{Range \& Domain \& Co-Appearance (RDC).} The negative impact of the co-appearance module carried over into the combination of type, domain, and co-appearance, resulting in decreased performance across all observed metrics, as well as increased execution time and failure rate in both datasets.

\textbf{Range \& Domain \& Co-Appearance based on precision ($RDC_p$).} For this experiment, we tested to activate the co-appearance module only on the relations that the model was able to identify with precision 100 \% in the Range \& Domain experiment. This is meaningful because 100 \% precision means that whenever the model spot this relation identifies it correctly. In order to achieve this it was calculated a per class score on validation set. Later it was activated manually only on these relations spotted by the LLM. In the R1 Dataset micro\_f1 score was increased by 1\% indicating that this approach works. Unfortunately in the R2 dataset this approach failed to improve the results, but compared to the approach with the full co-appearance micro\_f1 was increased by 15\%.

    

\begin{table}[t]
    \small
  \caption{CPA on R1 dataset. For approaches, D: Domain, R: Range, C: Co-App, RD: Range \& Domain, RDC: Range \& Domain \& Co-App, $RDC_p$: Range \& Domain \& Co-App(prec=1). \\For LLMs, L: llama3.3:70b-instruct-q2\_K, G: gemma3:12b, Q32: qwen2.5:32b-instruct-q3\_K\_L, Q72: qwen2.5:72b-instruct-q2\_K, R1: deepseek-r1:14b, $\Phi$: phi4:14b. }
  \label{tab:R1}
  \begin{tabular}{ccccccc}
    \toprule
    Approach & Macro\_F1 & Micro\_F1 & P & R & Time(s) & LLM\\
    \midrule
    Kepler-aSI ~\cite{DBLP:conf/semweb/BaazouziKF23} & - & 0.235 & 0.230 & - & -  & - \\
    \hdashline
    Base & 0.346 & 0.435 & 0.483 & 0.363 & 9.6 & L\\
    Base & 0.638 & 0.703 & 0.687 & 0.649 & 11.3 & Q72\\
    Base & 0.309 & 0.372 & 0.440 & 0.310 & \textbf{1.3} & G\\
    Base & 0.313 & 0.381 & 0.455 & 0.288 & 10.9 & R1\\
    Base & 0.202 & 0.453 & 0.249 & 0.206 & 6.46 & $\Phi$\\
    \hdashline
    Base & 0.634 & 0.687 & 0.706 & 0.662 & 3.9 & Q32\\
    R  & 0.679 & 0.694 & 0.748 & 0.681 & 3.8 & Q32\\
    D & 0.771 & 0.802 & 0.801 & 0.792 & 3.8 & Q32\\
    C & 0.054 & 0.232 & 0.092 & 0.051 & 4.1 & Q32\\
    RD & 0.792 & 0.812 & 0.834 & 0.801 & 3.8 & Q32\\
    RDC & 0.770 & 0.796 & 0.803 & 0.769 & 3.9 & Q32\\
    $RDC_p$ & \textbf{0.802} & \textbf{0.820} & \textbf{0.845} & \textbf{0.810} & 3.7 & Q32\\
    \hdashline
    $RDC_p$ & 0.441 & 0.575 & 0.544 & 0.428 & 1.6 & G\\
    $RDC_p$ & 0.185 & 0.339 & 0.248 & 0.179 & 7.5 & R1\\
    $RDC_p$ & 0.227 & 0.566 & 0.269 & 0.216 & 11.0 & $\Phi$\\
    
    \bottomrule
  \end{tabular}
\end{table}


\begin{table}[t]
    \small
  \caption{CPA on R2 dataset. For approaches, D: Domain, R: Range, C: Co-App, RD: Range \& Domain, RDC: Range \& Domain \& Co-App, $RDC_p$: Range \& Domain \& Co-App(prec=1). \\For LLMs, G: gemma3:12b, Q32: qwen2.5:32b-instruct-q3\_K\_L, R1: deepseek-r1:14b, $\Phi$: phi4:14b.}
  \label{tab:R2}
  \begin{tabular}{ccccccc}
    \toprule
    Approach & Macro\_F1 & Micro\_F1 & P & R & Time(s) & LLM\\
    \midrule
    TSOTSA ~\cite{DBLP:conf/semweb/JiomekongMTC23} & - & 0.235 & 0.434 & - & -  & -\\
    Anu & - & 0.623 & 0.791 & - & -  & -\\
    TorchicTab ~\cite{DBLP:conf/semweb/DasoulasYDD23} & - & \textbf{0.871} & \textbf{0.880} & - & - & -\\
    MUT2KG ~\cite{DBLP:conf/semweb/MehryarC23} & - & 0.793 & 0.848 & - & -  & -\\
    DREIFLUSS ~\cite{DBLP:conf/semweb/ParmarA23} & - & 0.174 & 0.320 & - & - & -\\
    \hdashline
    Base & 0.559 & 0.630 & 0.655 & 0.575 & 4.2 & Q32\\
    R  & 0.542 & 0.608 & 0.622 & 0.565 & 4.1 & Q32\\
    D & 0.704 & 0.739 & 0.756 & 0.717 & 4.0 & Q32\\
    C & 0.026 & 0.146 & 0.042 & 0.026 & 4.3 & Q32\\
    RD & \textbf{0.756} & \textbf{0.776} & \textbf{0.797} & \textbf{0.763} & \textbf{3.9} & Q32\\
    RDC & 0.478 & 0.654 & 0.525 & 0.474 & 4.7 & Q32\\
    $RDC_p$ & 0.685 & 0.749 & 0.734 & 0.695 & 4.1 & Q32\\
    \hdashline
    RD & - & - & - & - & - & G\\
    RD & 0.137 & 0.296 & 0.184 & 0.132 & 7.9 & R1\\
    RD & 0.200 & 0.548 & 0.239 & 0.192 & 27.5 & $\Phi$\\
    \bottomrule
  \end{tabular}
\end{table}

\begin{table}[t]
    \small
  \caption{Ablation study on prompts. 
  RD: Range \& Domain, $RDC_p$: Range \& Domain \& Co-App (precision=1),
  E: Example, E+COT: Example + COT.}
  \label{tab:prompts}
  \begin{tabular}{ccccccc}
    \toprule
    Approach & Macro\_F1 & Micro\_F1 & P & R & Time(s) & Without \\
    \midrule
    \multicolumn{7}{l}{\textbf{Dataset: R1}} \\
    $RDC_p$ & 0.766 & 0.799 & 0.816 & 0.773 & 4.0 & ALL\\
    $RDC_p$ & \textbf{0.813} & \textbf{0.835} & 0.844 & \textbf{0.818} & 3.7 & COT\\
    $RDC_p$ & 0.786 & 0.809 & 0.839 & 0.794 & 3.7 & Role\\
    $RDC_p$ & 0.805 & 0.826 & \textbf{0.845} & 0.816 & \textbf{3.7} & E\\
    $RDC_p$ & 0.782 & 0.796 & 0.834 & 0.791 & 3.7 & E+COT\\
    $RDC_p$ & 0.802 & 0.820 & \textbf{0.845} & 0.810 & 3.7 & -\\
    \hdashline
    \multicolumn{7}{l}{\textbf{Dataset: R2}} \\
    RD & 0.728 & 0.736 & \textbf{0.798} & 0.729 & \textbf{3.8} & ALL\\
    RD & 0.691 & 0.729 & 0.755 & 0.698 & 4.0 & COT\\
    RD & 0.722 & 0.763 & 0.775 & 0.729 & 3.9 & Role\\
    RD & 0.688 & 0.718 & 0.758 & 0.697 & 3.9 & E\\
    RD & 0.711 & 0.724 & 0.786 & 0.716 & 3.9 & E+COT\\\
    RD & \textbf{0.756} & \textbf{0.776} & 0.797 & \textbf{0.763} & 3.9 & -\\  
    \bottomrule
  \end{tabular}
\end{table}

 \begin{figure}
     \centering
     \includegraphics[width=\linewidth]{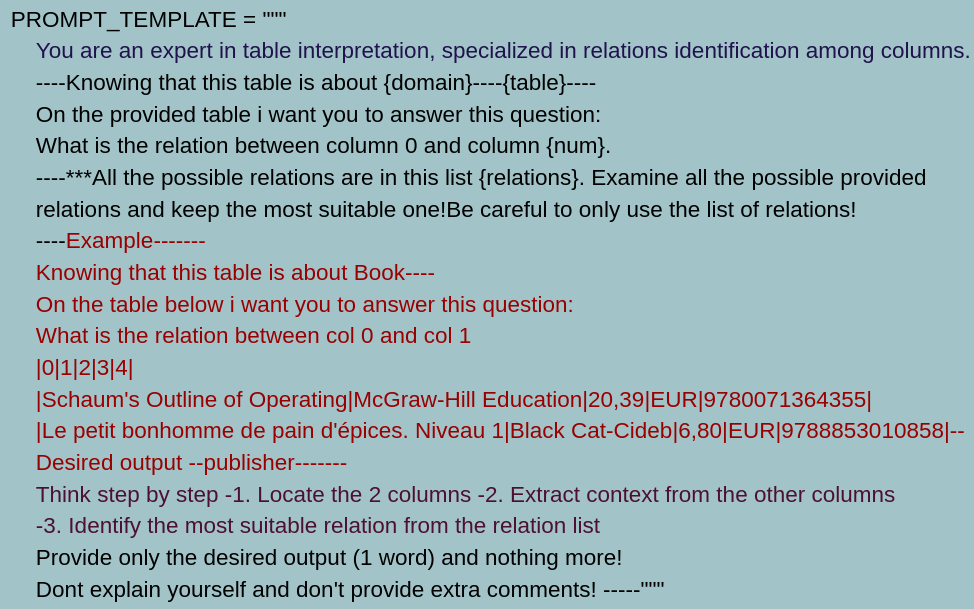}
     \caption{The main prompt that we used in this work, with its different parts highlighted in different colors (see Prompt Ablation).}
     \label{fig:prompt}
 \end{figure}

\textbf{Prompt Ablation.} On the R1 dataset, unexpectedly, the best micro\_F1 score resulted from the exclusion of CoT. Excluding the example also had a positive effect. Although this was initially considered unexpected, it can be explained by the fact that R1 is easier than R2 and the additional information may have confused the model. An overall improvement of 4.5\% in the micro\_F1 score was observed using prompt engineering.
On the R2 dataset the prompt with role, example and COT incorporation achieved the best results. Each prompt part exclusion had negative impact as well. In total using prompt engineering the micro\_f1 score was increased by 5.4\%. 
Although LLMs are improving in zero-shot performance, prompt engineering remains an easy way to boost results. However, it should be validated, as incorporating a seemingly beneficial component into the prompt may, in fact, have a negative impact.

\textbf{Comparison of LLMs.} Different models were tested under various quantization levels and parameters on both the base and final setups. On R1, DeepSeek and Phi-4 struggled with structuring, activating the structuring component, which led to higher execution time and failure rates. Gemma3 performed better in terms of execution time but was still less competitive than the proposed model. On R2, results declined across all models, indicating higher dataset complexity. While Phi-4 outperformed DeepSeek in output quality, it required over four times higher execution time. Gemma3 failed to complete the dataset due to frequent unexpected token errors, leaving its results blank.

\section{Conclusion}
In this work, a hybrid LLM-based and statistical analysis-based approach was explored for the CPA task. The employed LLMs varied in parameter size and quantization levels. Three main components were explored for the reduction of the search space of relationships: Domain, Range and Co-appearance. Each component reduces the relations based on the table's domain, the primitive type of the column, and the co-occurring relations, respectively. Finally, an ablation study on the constructed prompt is conducted which showcases the importance of the correct prompt utilization. The combination of domain and range approaches proved to be competitive with state-of-the-art methods in SemTab.

 In the future, we are planning to enhance the co-appearance methodology by selecting, as the initial point for the LLM, the column with the fewest possible relations, in order to reduce error propagation. Additionally, we will explore an alternative approach for classifying each column using pre-trained embeddings. In this approach, the embedding of each row in the column of interest would be aggregated, and the resulting vector would then be compared to the pre-trained embeddings of each possible relation. The closest match in the embedding space would be selected, ideally capturing the correct semantic meaning, or even extract the top k candidates with the above approach and parse them to the LLM. 

\begin{acks}
This work has received funding from the EU Horizon Europe research and innovation programme
through the GANNDALF project (GA No.
101084265) and the Ceasefire project (GA no. 101073876).
\end{acks}


\balance

\bibliographystyle{ACM-Reference-Format}
\bibliography{sample}

\end{document}